\documentclass[11pt]{article}

\usepackage[]{acl}

\usepackage{times}
\usepackage{latexsym}

\usepackage[T1]{fontenc}
\usepackage[utf8]{inputenc}
\usepackage{microtype}
\usepackage{inconsolata}
\usepackage{subcaption}
\usepackage{graphicx}
\usepackage{url}

\title{ACCIO: Table Understanding Enhanced \\
via Contrastive Learning with Aggregations}

\author{Whanhee Cho \\
  Kahlert School of Computing \\
  University of Utah \\
  \texttt{whanhee@cs.utah.edu} \\}

\begin{document}
\maketitle
\begin{abstract}
  The attention to table understanding using recent natural language models has
  been growing. However, most related works tend to focus on learning the
  structure of the table directly. Just as humans improve their understanding of
  sentences by comparing them, they can also enhance their understanding by
  comparing tables. With this idea, in this paper, we introduce \textbf{ACCIO}, t\textbf{A}ble understanding enhan\textbf{C}ed via \textbf{C}ontrastive
  learn\textbf{I}ng with aggregati\textbf{O}ns, a novel approach to enhancing
  table understanding by contrasting original tables with their pivot summaries
  through contrastive learning. ACCIO trains an encoder to bring these table
  pairs closer together. Through validation via column type annotation, ACCIO
  achieves competitive performance with a macro F1 score of 91.1 compared to
  state-of-the-art methods. This work represents the first attempt to utilize
  pairs of tables for table embedding, promising significant advancements in
  table comprehension. Our code is available at \url{https://github.com/whnhch/ACCIO/}.
\end{abstract}

\section{Introduction}

\begin{table}[t]
    \centering
    \captionsetup{skip=0pt}
    \begin{subtable}{\linewidth}
        \centering
        \begin{tabular}{lll}\hline
            \textbf{Year} & \textbf{Month} & \textbf{Passengers} \\\hline
            1949          & January         & 112                 \\
            1949          & February        & 118                 \\
            1949          & March           & 132                 \\
            1949          & April           & 129                 \\
            1949          & May             & 121                \\\hline
        \end{tabular}
        \caption{}
        \label{tab:sub_data}
    \end{subtable}
    \vspace{0.6cm} %
    \begin{subtable}{\linewidth}
        \centering
        \begin{tabular}{llll}\hline
            \textbf{Month}  & \textbf{1949} & \textbf{1950} & \textbf{1951} \\\hline
            \textbf{April}  & 129           & 135           & 163           \\
            \textbf{August} & 148           & 170           & 199           \\
            \textbf{December} & 118          & 140           & 166           \\
            \textbf{February} & 118          & 126           & 150           \\
            \textbf{January} & 112           & 115           & 145     \\\hline     
        \end{tabular}
        \caption{}
        \label{tab:sub_pivot}
    \end{subtable}
    \caption{Passenger data and pivot table. (a) is the original tabular data containing year and month passenger attributes. (b) is a pivot table with a user's query of ``the average number of passengers by month and year''}
    \label{tab:data_and_pivot}
\end{table}

Leveraging the success of natural language processing techniques, the
comprehension of tables has also significantly grown. Many works related to
understanding tables have been helpful in applications, such as column type
annotation, joining relation databases from data lakes, table to visualization,
and table normalization. These efforts typically analyze tables based on the
structure of the table, column relationships, or entity associations.

However, to our knowledge, no prior research has been aimed at enhancing table
understanding by comparing two tables. In the realm of sentence embeddings,
success has been achieved by comparing sentences using techniques like
SBERT~\cite{sbert} or SimCSE~\cite{simcse}, leveraging natural language
inference (NLI) datasets~\cite{snli, mnli}. These datasets typically contain
triplets consisting of a premise, an entailment, and a contradiction. The
entailment sentence can be logically inferred from the premise, while the
contradiction sentence directly contradicts the premise. Previous
studies~\cite{sbert, simcse} have introduced methods such as closing the
entailment sentence and premise and contrasting the premise with the contradiction
sentence, ultimately leading to high-quality sentence embeddings.

Therefore, in this paper, we exploit the notion that premise and entailment
sentences should be closely related by leveraging pivot tables and the original
table as the tables that should be conceptually close. Pivot tables are
summaries of tables using aggregation by user-defined parameters such as index,
column, value, and aggregation function. For instance, 
Table~\ref{tab:sub_data} represents the original tabular data consisting of
year, month, and passengers. Users typically analyze or summarize tables by
using pivot tables derived from such data. For example, when a user wants to
determine ``the average number of passengers by month and year,'' they can
obtain a pivot result like Table~\ref{tab:sub_pivot}. 

We present a novel approach to table understanding, called \textbf{ACCIO},
t\textbf{A}ble understanding enhan\textbf{C}ed via \textbf{C}ontrastive
learn\textbf{I}ng with aggregati\textbf{O}ns. By training an encoder with
original data and its corresponding pivot table using contrastive learning, we
aim to bring them closer together. It's worth highlighting that this paper marks
the first attempt to utilize a pair of tables for table embedding.

We validate our training method through a downstream task known as column type
annotation. This task is commonly used to evaluate the quality of table
embeddings by predicting the types of given columns. The performances indicate
that our approach achieves comparable performance in terms of macro F1 score
91.1 for column type annotation compared to state-of-the-art baselines.
\section{Related Works}
\subsection*{Table Understanding}
Table-GPT~\cite{tablegpt} is a generative pre-trained model (GPT) tailored for
tabular data. It trains GPT through instruction tuning across 14 types of table
tasks, each comprising 1000 template instances and augmented instructions and
tables. Table-GPT presents only one column and candidate column type from which
the model can choose. Another GPT approach relies solely on prompts to obtain
column-type annotations, with prompts consisting of partial
tables~\cite{annotchatgpt}. 

Doduo~\cite{doduo} annotates columns based on each column and relations between
columns using contextualized column embeddings and learning the tables from
column type annotations and relations. TURL~\cite{turl} focuses on entities in
tables by leveraging additional metadata such as table titles and captions from
Wikipedia tables, pre-training encoder with masked entity loss. Moreover,
Watchhog~\cite{watchog} employs contrastive learning within tables using a
self-supervised learning approach. It augments cells, rows, and columns in
various ways separately training each column. As mentioned earlier, with the
exception of using GPT, all these works attempt to incorporate additional
mechanisms to directly understand the structure of tabular data. In contrast,
our work simplifies the training method by solely contrasting two tabular
datasets.

\subsection*{Contrastive Learning}
Contrastive learning has emerged as one of the most popular methods for data embeddings in the field of deep learning~\cite{simclr, simcse, supcon}. This approach aims to minimize distances between similar data samples while maximizing distances between dissimilar ones. Essentially, a model is trained to bring positive examples (similar data) closer together and push negative examples (distinct data) farther apart. This mechanism, which doesn't require manually annotated labels, proves to be efficient in learning from datasets containing positive and negative examples. In the field of table understanding, Watchog~\cite{watchog} has already employed contrastive learning, but it's generally applied within a single table augmented from a single source. In contrast, ACCIO leverages two distinct tables, recognizing that one table can be inferred from the other in various ways, offering more diverse learning scenarios compared to augmentation methods.

\section{Methodoloy} 

\begin{figure*}[t]
    \centering
    \includegraphics[width=\textwidth]{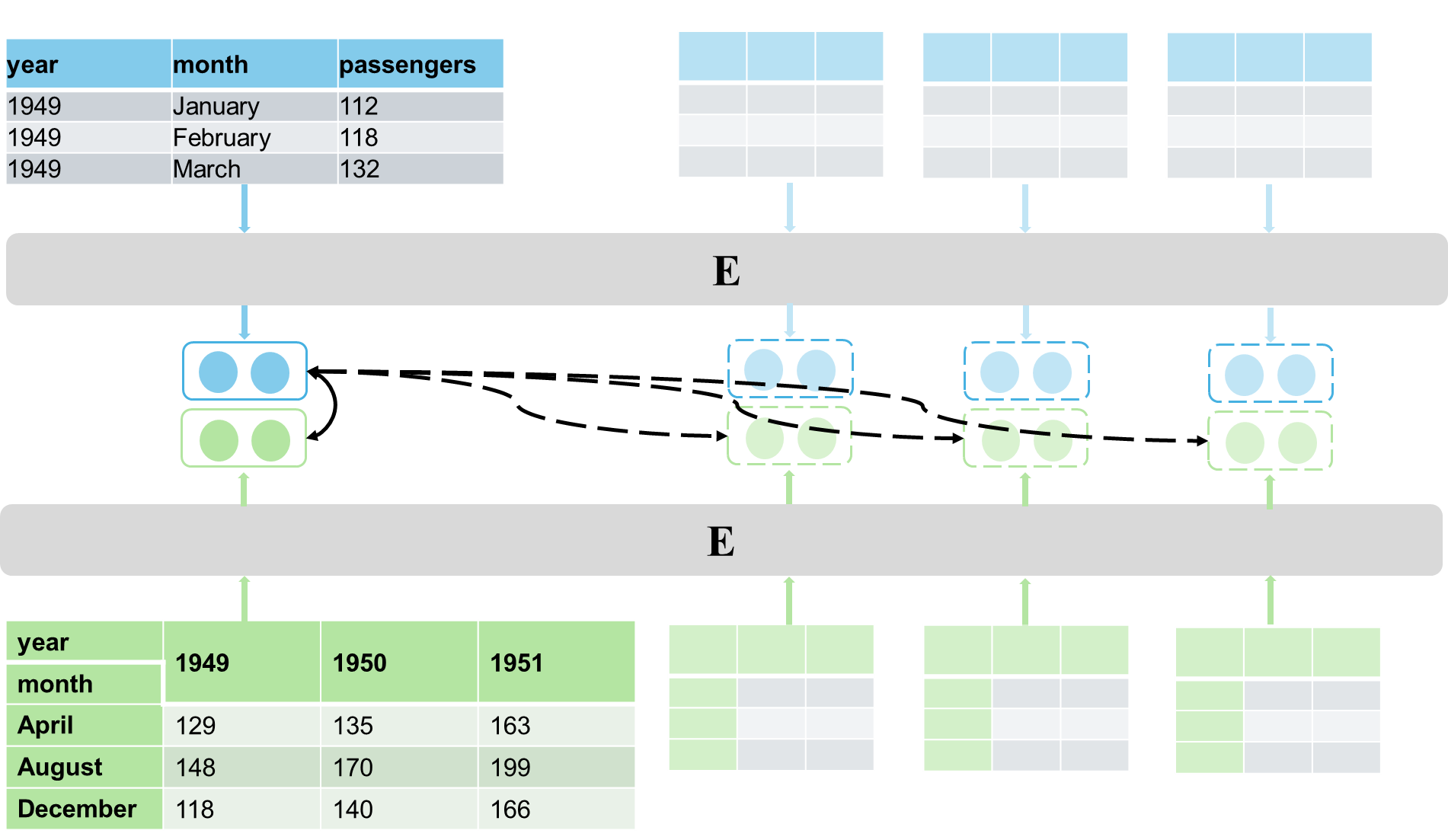} 
    \caption{ACCIO training method overview. ACCIO training involves annotating
    original tabular data (in blue) and corresponding pivot tables (in green).
    The solid lines represent positive pairs, where we aim to make the
    embeddings closer. In contrast, the dotted lines represent in-batch negative
    pairs, where we aim to make the embeddings farther apart. $E$ refers to the
    encoder.}
    \label{fig:accio}
\end{figure*}

\subsection{Serialization} 
Various serialization methods exist for tabular data. Some approaches embed data
column-wise and include a [\textit{CLS}] token for each column to facilitate
understanding of the data structure using appropriate learning
methods~\cite{doduo, annotating, watchog}. Table-GPT parses the data using
special characters like "|" to indicate to models that it represents a value in
tabular data~\cite{tablegpt}. TURL~\cite{turl} specifies types such as title,
caption, entity, or position embeddings. However, the majority of tokenization
methods primarily focus on understanding the structure of tabular data.

In our approach, we simplify table tokenization to prove the impact of
contrastive learning between data and its pivot tables by linearizing the table
column-wise. We serialize the table as shown in
Equation~\ref{eq:column_serialization}, where $h_i$ represents the $i$th column
header and $v_{ij}$ represents the value of the $i$th column and the $j$th row.
The header is placed at the beginning, and the column values are linearized with
spacing. The serialized table is then collected to form the serialized table
representation as shown in Equation~\ref{eq:table_serialization}, where $N$ is
the maximum number of columns. In this serialization, the table is appended with
\textit{[CLS]} at the beginning, and the columns are separated by \textit{[SEP]}.

\begin{equation}
    c_{i}=h_{i} \, v_{i1} \, v_{i2} \ldots v_{ij} \ldots
    \label{eq:column_serialization}
\end{equation}
\begin{equation}
    T=\textit{[CLS]} \, c_{1} \, \textit{[SEP]} \, c_{2} \, \textit{[SEP]}\ldots c_{N} \, \textit{[SEP]}
    \label{eq:table_serialization}
\end{equation}

\subsection{ACCIO}
Figure~\ref{fig:accio} shows the overall contrastive learning between tabular
data and pivot tables. We make embeddings closer between tabular data and its
pivot tables and farther between the tabular data and the other pivot tables. We
categorize pairs that need to be closer as "positive pairs" and those that need
to be farther apart as "negative pairs". Specifically, in this work, we only
consider negative pairs within the batch, so we refer to them as "in-batch
negative pairs".

We employ $h_{i}$, the average output of last hidden state of transformer-based
encoder model~\cite{bert}, as done in prior works~\cite{turl, watchog, doduo},
where the input is a table $T$, $i\in \{1,...,N\}$ referred in
equation~\ref{eq:table_serialization}. Each positive and in-batch negative pair
of the random columns could be represented as $(h_{i}, h_{i}^{+})$, and $(h_{i},
h_{j}^{+})$, where $i\neq j$. Then, we calculate distances in pairs with a
similarity function, $sim$, for here, $sim$ is cosine similarity.
Equation~\ref{eq:nt-xent} shows our loss function where $\tau$ is the
temperature hyperparameter. 

\begin{equation}
l_{i}=-\log \frac{e^{sim(h_{i},h_{i}^{+})/\tau}}{\Sigma_{j=1} e^{sim(h_{i},h_{j}^{+})/\tau}}
\label{eq:nt-xent}
\end{equation}
\section{Result}
In this section, we present the experiment environment of ACCIO and the
performance of the downstream task, column type annotation. We used
BERT-base-uncased from huggingface~\footnote{BERT-base-uncased:
\url{huggingface.co/google-bert/bert-base-uncased}}. NVIDIA RTX 6000 Ada 48GB. 

\subsection{Dataset}
For training ACCIO, we utilize pairs of tabular data and pivot table datasets
obtained from Auto-Suggest~\cite{autosuggest}. This dataset comprises 17,189
tables along with the corresponding pivot table parameters for the Pandas
function pivot table~\footnote{Pandas pivot table:
\url{pandas.pydata.org/docs/reference/api/pandas.pivot_table.html}}. Initially,
we generated pivot tables to construct our dataset consisting of tabular data
and their corresponding pivot tables. Additionally, since we linearize tabular
data, simply storing the table and setting a maximum length would not account
for diverse rows. Therefore, we set the number of columns to 10 and the number
of rows to 10.

For the column type annotation task to validate our work, we utilize the Viznet
dataset, which contains 119,360 columns preprocessed by Watchog~\cite{watchog}.
Watchog categorized the types of columns into 78 semantic types. Similar to
prior work, we conduct experiments on multi-classification tasks by attaching a
linear layer to the encoder and employing 5-fold cross-validation.

\subsection{Contrastive Learning}
\begin{figure}[h]
    \centering
    \includegraphics[width=\linewidth]{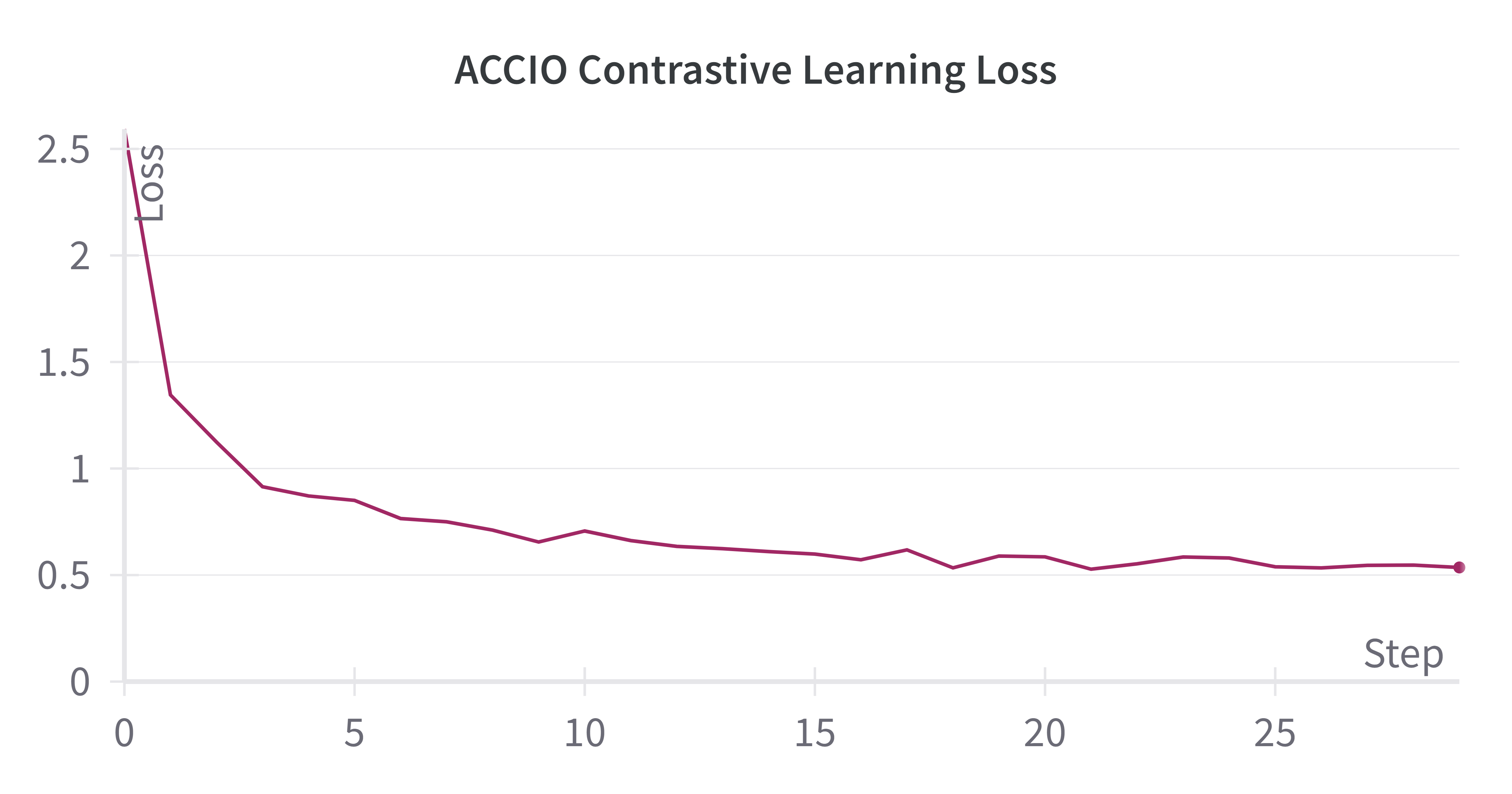} 
    \caption{ACCIO contrastive learning loss.}
    \label{fig:accio_training_loss}
  \end{figure}

We trained ACCIO with a learning rate of 1e-5 using the Adam optimizer, a batch
size of 64, the maximum sequence length of 256, and 5 epochs, with $\tau$ set to
0.05. Our observations indicate that the training loss converges as shown in
Figure~\ref{fig:accio_training_loss}. Therfore, we concluded that the trained
model generalized enough from the contrastive learning.
  
\subsection{Column Type Annotation}
We appended a linear layer to the trained encoder, with a size of (768, 78) to
match the 78 types and the average dimension of the last hidden state, which is
768. We trained the model with cross-entropy loss. Training the model with a
maximum sequence length of 256, a batch size of 128, and 4 epochs, with a
learning rate of 1e-4 using the Adam optimizer, resulted in our model achieving
a macro F1 score of 91.1 and a micro F1 score of 77.3, as shown in
Table~\ref{tab:result}.

\begin{table}[h]
  \centering
    \begin{tabular}{lll} \hline
    Method   & Micro F1 & Macro F1 \\ \hline
    Sherlock$^\dagger$ & 86.7     & 69.2     \\
    SATO$^\dagger$     & 88.4     & 75.6     \\
    Doduo$^\dagger$    & 94.3     & 84.6     \\
    Starmie$^\dagger$  & 94.0    & 83.6    \\
    Watchog$^\dagger$  & \textbf{95.0}    & \textbf{85.6}    \\ \hline
    ACCIO   & 91.1        &  77.3     \\ \hline
    \end{tabular}
    \caption{ACCIO performances on VizNet. $^\dagger$: results from \citet{watchog}.}
    \label{tab:result}
\end{table}

ACCIO exhibits superior performance on micro F1 compared to macro F1 when
contrasted with the state-of-the-art model, Watchog. This variance may stem from
the unbalanced nature of the Viznet dataset, which our method appears to be
sensitive to.
\section{Future Work}
\subsubsection*{Quantity and Quality of Pivot Tables}
Our method is limited by the number of contrastive learning training instances.
With a dataset of 17,189 tables, it may struggle to generalize tables
effectively. Wathhog leverages data augmentation methods within tables which doubles the training data. Additionally, due to the nature of pivot tables, some tables
contain numerous null values, which poses a challenge for our approach. While
our method aims to enhance table understanding by contrasting similar tables,
the presence of null values can confuse the model, particularly when headers
align, but most values are null. In the future, we could consider creating pivot
tables from scratch and preprocessing semi-empty pivot tables to mitigate
potential distractions to our work. 

\subsubsection*{Structure Understanding}
Since we simplified table tokenization and did not incorporate additional mechanisms for understanding the structure, we cannot guarantee that our model can fully comprehend table structures based solely on column type annotation tasks. In the future, we should include other downstream tasks, such as column relation analysis, to enhance the model's understanding of table structures.

\section{Conclusion}
ACCIO demonstrates competitive performance in table understanding by leveraging contrastive learning to compare tabular data and its pivot tables. This method marks the first attempt to enhance table comprehension through table comparisons. While it may not surpass state-of-the-art performance, this straightforward approach provides valuable insights for future research in the field.

\bibliography{paper}
\bibliographystyle{acl_natbib}

\end{document}